\pdfoutput=1

\documentclass[11pt]{article}
\usepackage[final]{EMNLP2023}
\usepackage{times}
\usepackage{subcaption}
\usepackage{latexsym}
\usepackage[T1]{fontenc}
\usepackage[utf8]{inputenc}
\usepackage{float}
\usepackage{microtype}
\usepackage{inconsolata}
\usepackage{url}
\usepackage{booktabs}
\usepackage{amsfonts}
\usepackage{nicefrac}
\usepackage{graphicx}
\usepackage{caption}
\usepackage{lineno}
\usepackage{placeins}
\usepackage{multirow}
\usepackage{amsmath}
\usepackage{amssymb}
\usepackage{longtable}
\usepackage{colortbl}
\usepackage{tabularx}
\usepackage{algorithm}
\usepackage{algorithmic}
\usepackage{xspace}

\title{AnchorAttention: Difference-Aware Sparse Attention with Stripe Granularity}

\author{
\textbf{Yu Zhang}$^{1,*}$, 
\textbf{Dong Guo}$^{3,*}$, 
\textbf{Fang Wu}$^{1}$, \\
\textbf{Lu Tang}$^{1}$,
\textbf{Guoliang Zhu}$^{4}$,
\textbf{Dian Ding}$^{2}$,
\textbf{Yiming Zhang}$^{1,2,\dagger}$ \\
$^1$Xiamen University, 
$^2$Shanghai Jiao Tong University,
$^3$Xi'an Jiaotong University, 
$^4$Chipltech \\
}

\begin{document}
\maketitle

\begin{abstract}
Large Language Models (LLMs) with extended context lengths face significant computational challenges during the pre-filling phase, primarily due to the quadratic complexity of self-attention. Existing methods typically employ dynamic pattern matching and block-sparse low-level implementations. However, their reliance on local information for pattern identification fails to capture global contexts, and the coarse granularity of blocks leads to persistent internal sparsity, resulting in suboptimal accuracy and efficiency. To address these limitations, we propose \textbf{AnchorAttention}, a difference-aware, dynamic sparse attention mechanism that efficiently identifies critical attention regions at a finer stripe granularity while adapting to global contextual information, achieving superior speed and accuracy. AnchorAttention comprises three key components: (1) \textbf{Pattern-based Anchor Computation}, leveraging the commonalities present across all inputs to rapidly compute a set of near-maximum scores as the anchor; (2) \textbf{Difference-aware Stripe Sparsity Identification}, performing difference-aware comparisons with the anchor to quickly obtain discrete coordinates of significant regions in a stripe-like sparsity pattern; (3) \textbf{Fine-grained Sparse Computation}, replacing the traditional contiguous KV block loading approach with simultaneous discrete KV position loading to maximize sparsity rates while preserving full hardware computational potential.
With its finer-grained sparsity strategy, \textbf{AnchorAttention} achieves higher sparsity rates at the same recall level, significantly reducing computation time. Compared to previous state-of-the-art methods, at a text length of 128k, it achieves a speedup of 1.44$\times$ while maintaining higher recall rates. 
Code is available at \url{https://github.com/yuzhouzhang9/Anchor-Attention}.

\end{abstract}

\renewcommand{\thefootnote}{}
\footnotetext{$^*$Equal contribution}
\footnotetext{$^\dagger$Corresponding author}

\section{Introduction}

Large Language Models (LLMs) have brought transformative advancements to numerous domains by enabling sophisticated natural language understanding and generation\cite{zhou2024largelanguagemodelsdisease,kaddour2023challengesapplicationslargelanguage,qin2024largelanguagemodelsmeet}.
However,
as the supported context lengths continue to increase,
the inference cost — particularly in the prefill phase — has become a major bottleneck. 
This is primarily due to the quadratic computational complexity of full-attention mechanisms with respect to sequence length,
which leads to significant efficiency issues in long-sequence inference tasks.

\begin{figure}[t]
    \centering
    \begin{subfigure}[b]{0.48\columnwidth}
        \includegraphics[width=\textwidth]{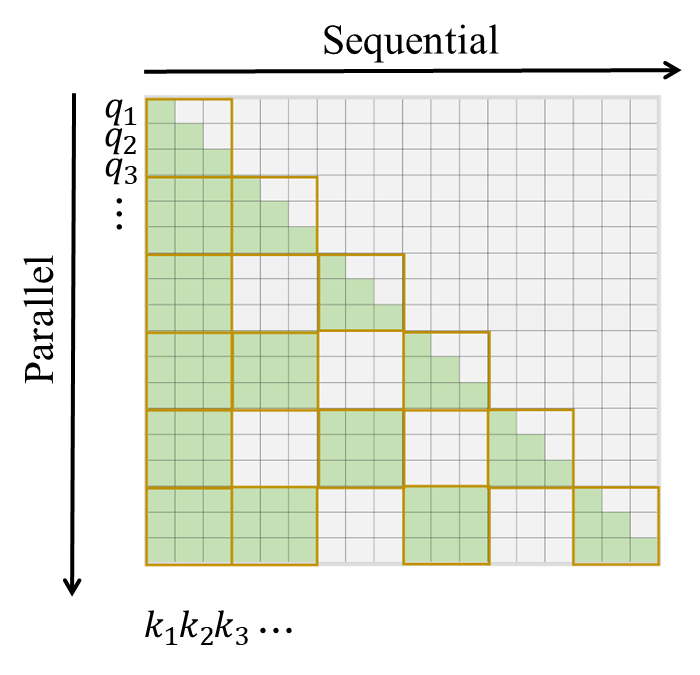}
        \caption{Block sparse}
    \end{subfigure}
    \hfill
    \begin{subfigure}[b]{0.48\columnwidth}
        \includegraphics[width=\textwidth]{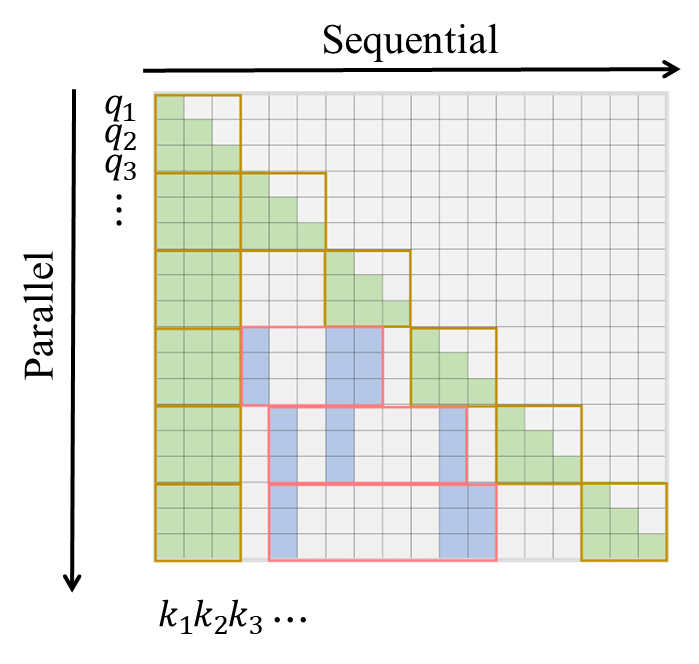}
        \caption{Stripe sparse}
    \end{subfigure}
    \caption{(a) Block-sparse pattern, with yellow regions indicating computed blocks; (b) Stripe-sparse pattern, with red regions showing computed areas, enabling higher sparsity by loading non-contiguous positions across multiple blocks.}
    \label{fig:block_stripe_comparison} 
\end{figure}

To mitigate the computational overhead during the prefill phase, FlashAttention~\cite{dao2022flashattention} leverages memory transfer disparities across hardware hierarchies and incorporates the online Softmax algorithm~\cite{milakov2018onlinenormalizercalculationsoftmax}, thereby significantly reducing transmission costs at the hardware level. Meanwhile, several studies~\cite{li2024snapkv,zhang2023ho,fu2024headsmatterheadlevelkv,yang2024pyramidinferpyramidkvcache} have revealed the inherent sparsity in attention mechanisms, demonstrating that retaining only a small subset of key-value (KV) pairs is sufficient to preserve model accuracy. However, these methods still rely on computing full attention scores to identify the retained KV subset and therefore do not reduce the runtime cost during the prefill phase. Recent efforts have attempted to exploit sparsity to optimize prefill computation. For example, StreamingLLM~\cite{xiao2024efficientstreaminglanguagemodels} introduces a sparse pattern that retains only local and initial positions during computation, significantly accelerating attention, but often missing essential information from intermediate content. Minference~\cite{jiang2024minference10acceleratingprefilling} proposes that attention patterns follow multiple coefficient modes and accelerate computation by applying offline-searched sparse configurations. However, its static design is not adaptive to diverse input patterns and often fails to select optimal configurations. FlexPrefill~\cite{lai2025flexprefillcontextawaresparseattention} improves upon this by dynamically selecting patterns online, yet its selection heavily depends on local information, limiting its generality. SpargeAttn~\cite{zhang2025spargeattnaccuratesparseattention} and X-Attention~\cite{xu2025xattentionblocksparseattention} attempt to identify informative blocks using similarity-based or diagonal priors. However, these designs primarily target general-purpose models and lack specialized mechanisms for language model characteristics, particularly in exploiting architectural properties like the \textit{attention sink}\cite{xiao2024efficientstreaminglanguagemodels} phenomenon. On the other hand, as shown in Fig.~\ref{fig:block_stripe_comparison}, current methods generally rely on coarse-grained block-level KV selection in attention computation, which is misaligned with the naturally fine-grained sparsity observed in attention maps, inevitably leading to redundant attention computations.

\begin{figure}[t]
    \centering
    \begin{subfigure}[b]{\columnwidth}
        \includegraphics[width=\textwidth]{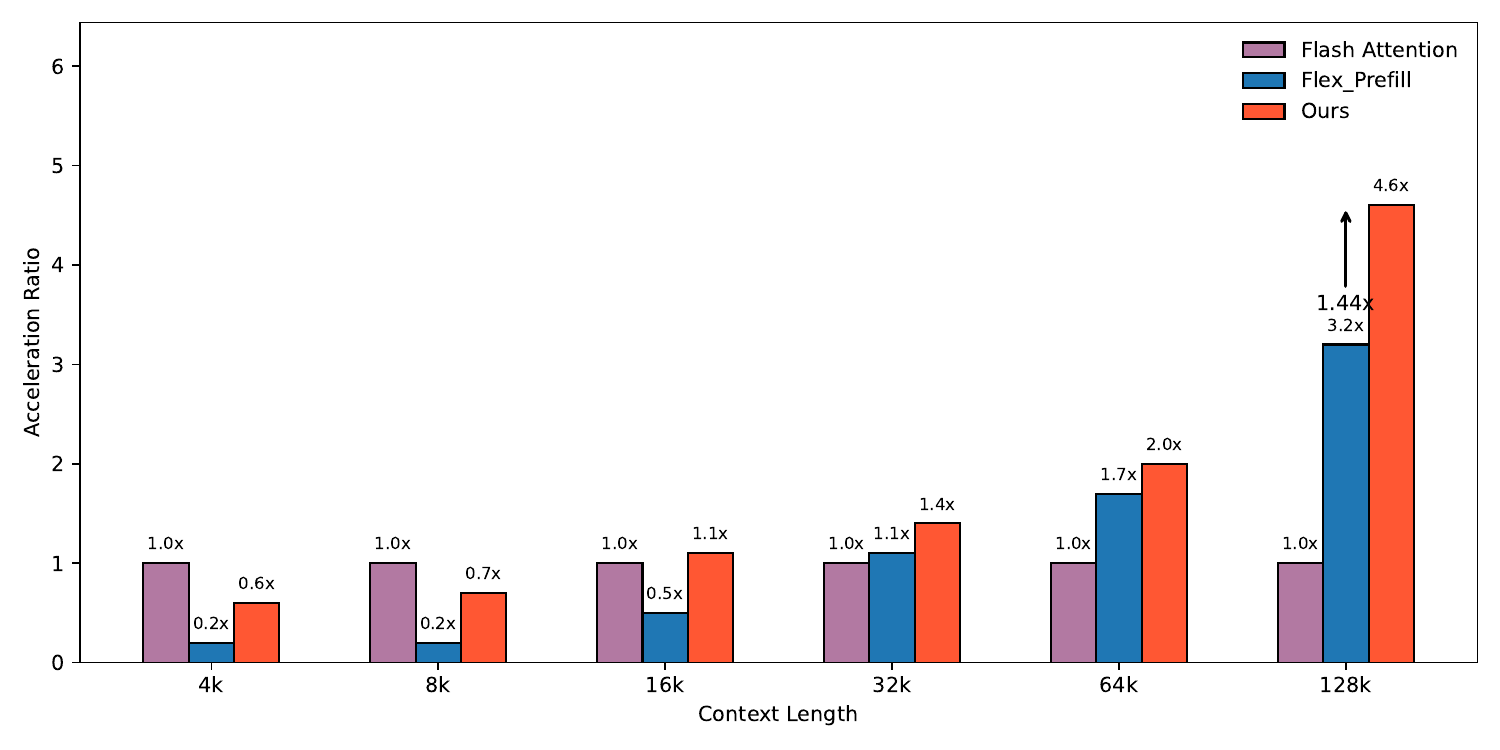}
    \end{subfigure}
    \caption{Acceleration of attention computation compared to FlashAttention.}
    \label{fig:block_stripe_comparison} 
\end{figure}

To address these challenges, we propose \textbf{AnchorAttention}, a difference-aware sparse attention strategy with stripe granularity. AnchorAttention introduces a sparsity mechanism centered around the concept of an \textbf{anchor}, inspired by the common structural patterns observed in attention distributions across all inputs. We observe that the maximum values after dot-product computations consistently emerge at initial or local window positions. We therefore extract the maximum score from these regions and designate it as the \textbf{anchor}. The importance of other positions is then determined by directly comparing their values against the anchor, effectively bypassing expensive sorting operations. In contrast to traditional block-level sparsity methods~\cite{zhang2025spargeattnaccuratesparseattention,xu2025xattentionblocksparseattention,lai2025flexprefillcontextawaresparseattention,jiang2024minference10acceleratingprefilling,yang2025lserveefficientlongsequencellm}, AnchorAttention adopts a more flexible \textit{stripe-level sparsity} strategy, reducing the identification granularity from coarse blocks to finer-grained stripes and enabling higher sparsity rates. During sparse computation, we further replace the conventional contiguous KV loading scheme with a discrete KV loading approach, which enhances recognition precision over block-based strategies while preserving parallel computation efficiency.
\textbf{AnchorAttention} comprises the following three steps:\textbf{Pattern-based Anchor Computation}: We observe that the distribution of the most significant values remains fixed and stable across various input transformations. We first compute these values and designate the obtained approximate maximum value as the anchor.\textbf{Difference-aware Stripe Sparsity Identification}: Compared to block sparsity, we adopt a finer-grained stripe sparsity approach. By performing dot-product computations between the compressed query and the full set of keys, we use direct comparisons with the anchor's difference to rapidly identify which keys and values are significant, avoiding costly sorting operations.\textbf{Fine-grained Sparse Computation}: We transition from block sparsity's continuous KV loading to discrete KV loading. During computation, we maintain block-based computations to maximize sparsity while preserving parallel computing capabilities.

We evaluate \textbf{AnchorAttention} on Llama-3.1-8B-Instruct~\cite{touvron2023llama} and Qwen2.5-7B-Instruct\cite{qwen2025qwen25technicalreport} across various context lengths. The benchmarks used include RULER~\cite{hsieh2024rulerwhatsrealcontext}, Needle In A Haystack~\cite{kamradt2023llmtest}, and Longbench~\cite{bai2024longbenchbilingualmultitaskbenchmark}.All of our experiments are conducted under context lengths up to 128k. Our goal is not to endlessly extend the context length while relying on simple-task performance as the evaluation metric but rather to approximate full attention with minimal computation. Therefore, we adopt recall as the primary evaluation metric. Under this criterion, our method surpasses the state-of-the-art FlexPrefill~\cite{lai2025flexprefillcontextawaresparseattention} in recall while achieving a 1.44$\times$ speedup. Compared to full KV FlashAttention~\cite{dao2022flashattention}, our method achieves a 4.6$\times$ speedup, significantly reducing attention computation time. The results demonstrate that \textbf{AnchorAttention} delivers substantial acceleration while preserving model accuracy.
\begin{figure}[htbp]
    \centering
    \begin{subfigure}[b]{0.48\columnwidth}
        \includegraphics[width=\textwidth,height=4cm,keepaspectratio]{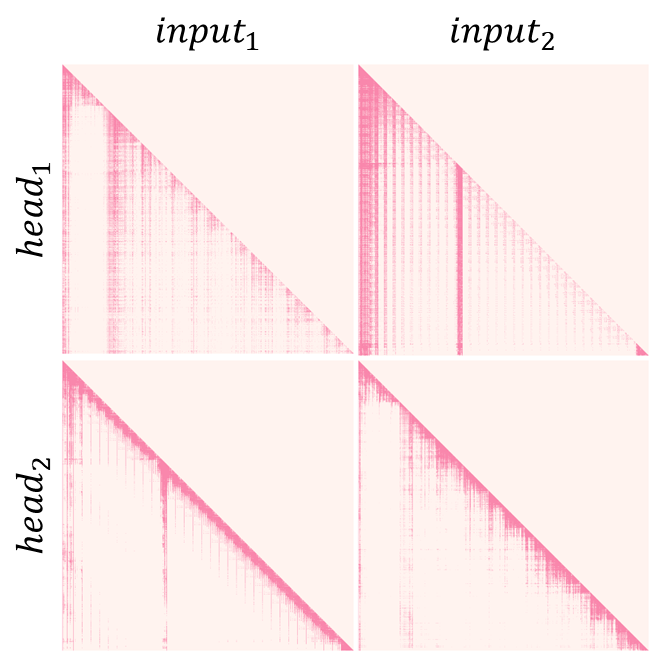}
        \caption{Heatmaps of different inputs}
        \label{fig:diff_model}
    \end{subfigure}
    \hfill
    \begin{subfigure}[b]{0.48\columnwidth}
        \includegraphics[width=\textwidth,height=4cm,keepaspectratio]{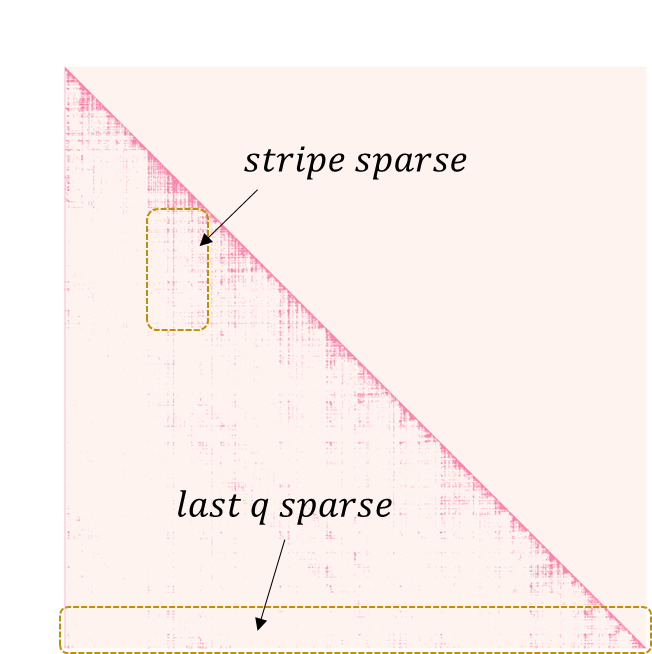}
        \caption{Stripe sparse and local information sparse}
        \label{fig:last_q}
    \end{subfigure}
    \caption{(a) Heatmaps vary significantly across different inputs. (b) Stripe sparse appears in specific attention maps, demonstrating that local information fails to capture the full attention distribution.}
    \label{fig:ab}
\end{figure}

\section{Analysis and Observation}
\label{Insight}

\subsection{Analysis}
\label{sec:sparsity_strategies}
In this section, we primarily analyze the impact of identification schemes~\label{sec:Identification_Schemes} and identification granularities~\label{sec:Granularities} on the final recall rate, elucidating how different identification approaches and granularities affect the output.

\begin{figure*}[htbp]
    \centering

    \begin{subfigure}[t]{0.32\textwidth}
        \centering
        \includegraphics[width=\textwidth]{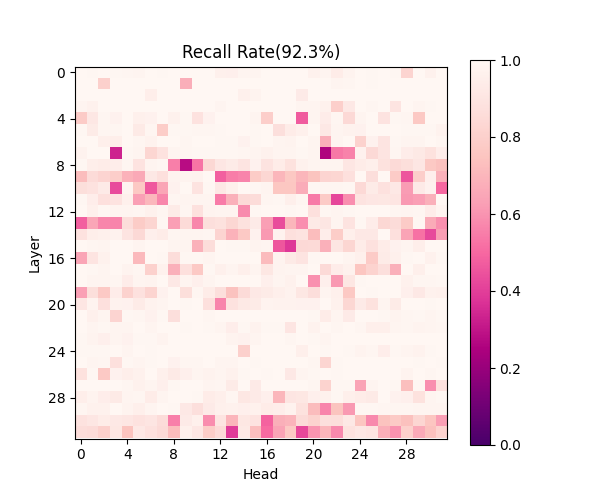}
        \caption{Top-K(4096)}
        \label{fig:heat_map_top_k}
    \end{subfigure}
    \hfill
    \begin{subfigure}[t]{0.32\textwidth}
        \centering
        \includegraphics[width=\textwidth]{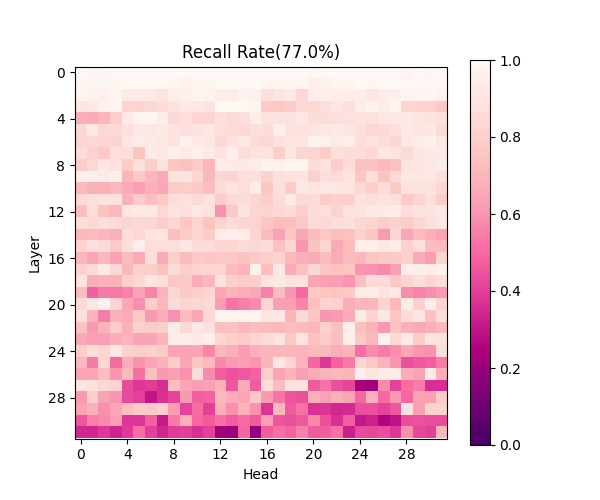}
        \caption{Top-CDF(0.95)}
        \label{fig:heat_map_top_cdf}
    \end{subfigure}
    \hfill
    \begin{subfigure}[t]{0.32\textwidth}
        \centering
        \includegraphics[width=\textwidth]{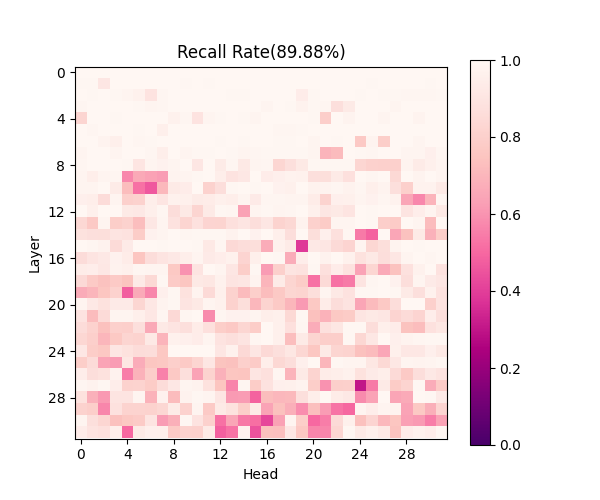}
        \caption{Difference-Aware(11)}
        \label{fig:heat_map_difference}
    \end{subfigure}
    \caption{
    Recall heatmaps of Sparsity Strategies using LLaMA-3.1-8B on the 128k Ruler\cite{hsieh2024rulerwhatsrealcontext} dataset, with average sparsity rates of 93.7\% (a), 96.4\% (b), and 94.1\% (c). Per-head sparsity rates are detailed in Appendix~\ref{app:Difference_method_sparisty}. Recall is defined as the percentage of attention values that are numerically equal between the current sparse attention and the full attention\cite{jiang2024minference10acceleratingprefilling}.
    }
    \label{fig:heatmaps}
\end{figure*}

\subsubsection{Performance of Different Identification Schemes in Sparsity Strategies}
\label{sec:dynamic_method_sparisty}
Previous work has widely adopted top-k\cite{xiao2024efficientstreaminglanguagemodels,li2024snapkv,holmes2024deepspeedfastgenhighthroughputtextgeneration,tang2024questqueryawaresparsityefficient,liu2024retrievalattentionacceleratinglongcontextllm} and top-cdf\cite{lai2025flexprefillcontextawaresparseattention} strategies to identify important positions in sparsity strategies. In the top-k strategy, the value of \( k \) is fixed. As shown in Figure~\ref{fig:heat_map_top_k}, this static selection of \( k \) can result in some heads having recall rates well below the target, prompting prior methods to assign different \( k \) values for different heads. However, such static \( k \) settings often perform poorly with dynamic inputs, as further detailed in Appendix~\ref{app:dynamic_method_sparisty}. To address this limitation, some methods employ the top-cdf strategy (see Figure~\ref{fig:heat_map_top_cdf}), which ensures each head meets the desired recall rate by computing cumulative attention scores. However, both approaches rely on sorting, incurring significant computational overhead. In contrast, the difference-aware strategy (see Figure~\ref{fig:heat_map_difference}) begins with a known maximum value and directly subtracts other values to obtain the differences. If the difference exceeds a predefined threshold, subsequent computations are skipped. This method eliminates the need for sorting operations and achieves performance comparable to that of top-cdf, while the maximum value, as discussed in Section~\ref{sec:attention_commonality}, can be obtained with minimal computational overhead.

\begin{table}[htbp]
    \centering
    \setlength{\tabcolsep}{4pt}
    \begin{tabular}{lcc}
        \toprule
        Method & Recall Rate & Sparsity Rate \\
        \midrule
        Block (Top-K=256) & 88.5\% & 56.3\% \\
        Stripe (Top-K=16384) & 91.2\% & 76.6\%  \\
        \bottomrule
    \end{tabular}
    \caption{Comparison of block and stripe granularity in sparsity strategies for LLaMA-3.1-8B on the 128k ruler\cite{hsieh2024rulerwhatsrealcontext} dataset.}
    \label{tab:granularity_comparison}
\end{table}

\subsubsection{Performance of Different Identification Granularities in Sparsity Strategies}
\label{sec:Granularities}
In Section~\ref{sec:dynamic_method_sparisty}, we systematically analyzed the impact of various strategies on the final recall rate. However, identifying these positions requires computing full attention scores, offering only limited acceleration for attention computation. Many prior methods rely on the underlying block‐sparse attention implementation, employing different block identification approaches. Yet, as discussed in Section~\ref{sec:attention_diversity}, not all elements within a block are equally significant; the heatmap often exhibits a stripe sparse. To address the issue of overly coarse block granularity, we simplify the block size to retain only the column dimension and set the row dimension to 1, which we term “stripe granularity.”

Through comparative experiments between stripe‐granularity strategies and traditional block‐sparse identification (block granularity(128,128) vs. stripe granularity(128,1)), we evaluated achievable sparsity rates while maintaining equivalent recall rates. As shown in Table~\ref{tab:granularity_comparison}, the stripe‐granularity approach achieves higher sparsity rates at comparable or higher recall thresholds. This finding offers an innovative, implementation‐level alternative to traditional block‐sparse solutions via stripe‐based sparsification.

\subsection{Observation}
\label{sec:Anchors_attention}

\subsubsection{Diversity of Sparse Attention Patterns}
\label{sec:attention_diversity}

Sparse attention patterns are prevalent in large language models, yet the sparsity distribution within a single attention head varies significantly due to input content\cite{lai2025flexprefillcontextawaresparseattention}. As shown in Figure~\ref{fig:diff_model}, different inputs yield distinct sparsity patterns, indicating that static pattern recognition cannot adapt to dynamic inputs,  \textbf{necessitating more flexible sparsity strategies}. 
Additionally, Figure~\ref{fig:last_q} shows that critical information often appears at a finer granularity, concentrating in only a few columns and forming a striped pattern in the heatmap. This phenomenon highlights that using block sparsity as the minimum granularity fails to fully leverage sparsity, \textbf{underscoring the need for finer-grained selection strategies}.

Moreover, Figure~\ref{fig:last_q} demonstrates that relying solely on the local information from the last query fails to reconstruct the full attention heatmap\cite{jiang2024minference10acceleratingprefilling,lai2025flexprefillcontextawaresparseattention}, as these stripes may vanish at the end, highlighting that local information lacks generalizability and \textbf{requiring broader positional data}.

\subsubsection{Commonality of Sparse Attention Patterns}
\label{sec:attention_commonality}

\begin{figure}[tbp]
    \centering
    \includegraphics[width=\columnwidth,keepaspectratio]{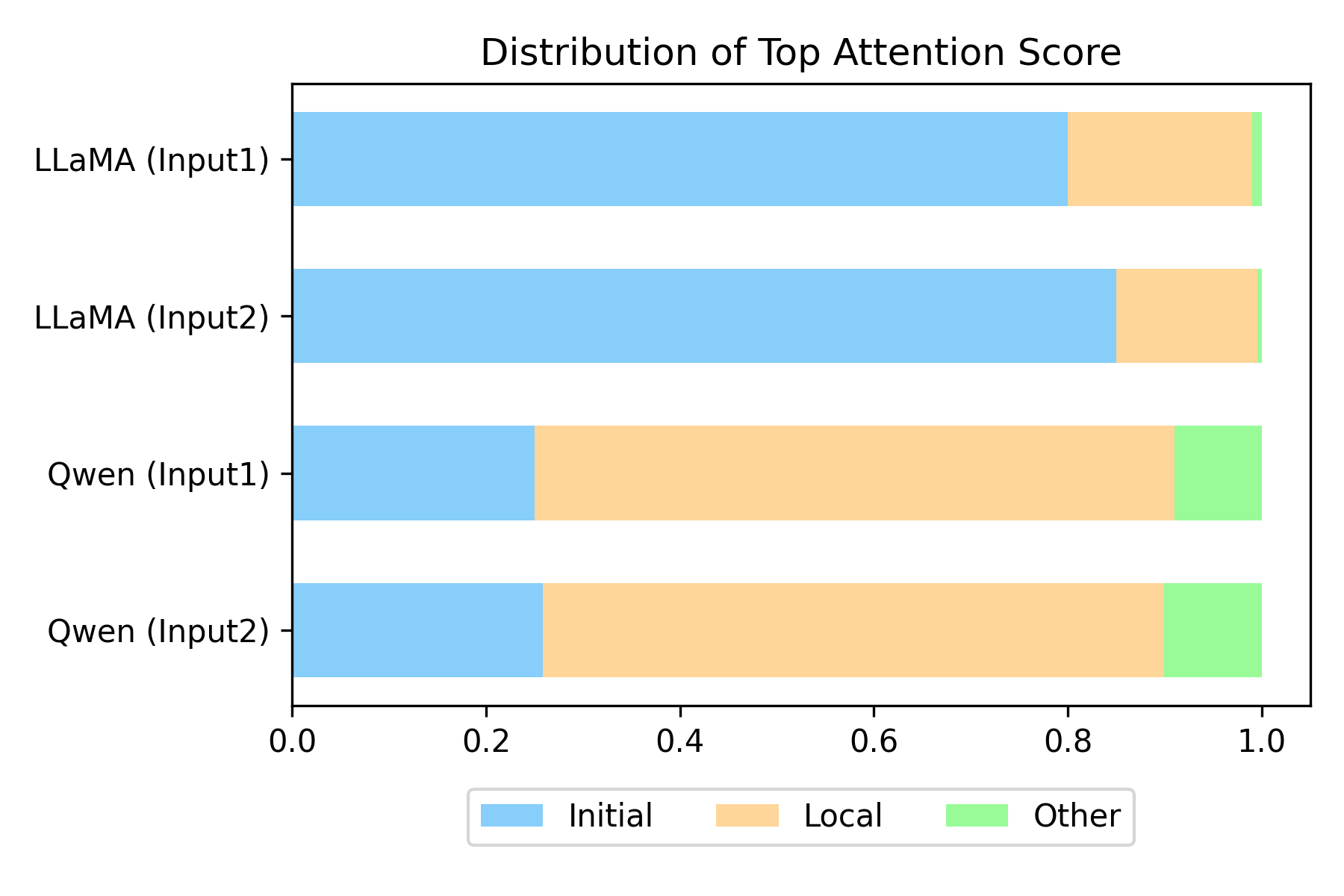}
    \caption{The distribution of maximum attention scores highlights the dominance of anchor positions.}
    \label{fig:anchor_att}
\end{figure}

Although sparsity patterns vary significantly across different models, certain consistent features remain prominent. As shown in Figures~\ref{fig:diff_model} and~\ref{fig:last_q}, the attention scores \textbf{at the local window positions and the initial token position are consistently critical}. We further analyze these positions in Figure~\ref{fig:anchor_att}, examining the first token and a local window of 128 tokens under a 128k context length. The results show that in the LLaMA~\cite{touvron2023llama} model, approximately 99\% of the highest attention scores are concentrated in these regions, whereas in the Qwen~\cite{qwen2025qwen25technicalreport} model, the proportion is around 90\%. Although prior works~\cite{xiao2024efficientstreaminglanguagemodels, jiang2024minference10acceleratingprefilling} have identified the importance of these positions and focused on preserving them, their potential for guiding the construction of broader sparsity structures remains underexplored. In contrast, we propose to define these high-impact positions as \textbf{anchor}, emphasizing their critical role in attention computation and their utility in precomputing and approximating the sparsity distribution of other positions.

\section{Method}
\label{sec:method}

In this section, we present \textbf{AnchorAttention}, a \textbf{difference-aware} and \textbf{stripe} sparse attention strategy. \textbf{AnchorAttention} consists of three key components:  
(1) \textbf{Pattern-based Anchor Computation},  
(2) \textbf{Difference-aware Stripe Sparsity Identification}, and  
(3) \textbf{Fine-Grained Sparse Computation}.
We implement all three strategies as kernel operations, as described in (4) \textbf{Kernel Optimization and Algorithm}.

\subsection{Pattern-based Anchor Computation}
\label{sec:anchor_computation}

As discussed in Section~\ref{sec:attention_commonality}, attention scores consistently exhibit prominent peaks in two specific regions: the initial token positions and the local window position. This structurally stable pattern motivates us to explicitly compute the attention scores at these positions and define the resulting maximum value as the \textbf{anchor}.

The anchor computation is highly efficient, as it requires only a small subset of the key. This enables us to approximate the maximum attention score at a very low computational cost, avoiding the need to compute the full attention. The computed anchor can then directly guide the selection of sparse attention patterns.

Formally, the anchor is computed as follows:
\begin{equation}
\boldsymbol{x}_{\text{a}} = \max\left( \frac{Q [K_{\text{init}}, K_{\text{w}}]^T}{\sqrt{d}}, \text{dim}=-1 \right)
\label{eq:anchor}
\end{equation}
where \( Q \) is the query matrix, \( [K_{\text{init}}, K_{\text{w}}] \) is the concatenated key vectors, \( K_{\text{init}} \) corresponds to the initial tokens, \( K_{\text{w}} \) corresponds to the local window, both selected as blocks for computation. The resulting \( \boldsymbol{x}_{\text{a}} \) is the highest score observed within the structurally important regions, which we define as the anchor, as detailed in Algorithm~\ref{alg:anchor_comp}.

\begin{table*}[t]
\centering
\resizebox{\textwidth}{!}{
\begin{tabular}{llccc|ccc|ccc|ccc|cc|cc|c}
\toprule
\multirow{2}{*}{\textbf{Models}} & \multirow{2}{*}{\textbf{Methods}} &
\multicolumn{3}{c|}{\textbf{Single-Document QA}} &
\multicolumn{3}{c|}{\textbf{Multi-Document QA}} &
\multicolumn{3}{c|}{\textbf{Summarization}} &
\multicolumn{3}{c|}{\textbf{Few-shot Learning}} &
\multicolumn{2}{c|}{\textbf{Synthetic}} &
\multicolumn{2}{c|}{\textbf{Code}} &
\multirow{2}{*}{\textbf{Avg.}} \\
& & NarrQA & Qasper & MF-en &
HotpotQA & 2Wiki & Musique &
GovRep & QMSum & MNews &
TREC & Trivia & SAMSum &
PCount & PR-en &
Lcc & RP-P & \\
\midrule
\multirow{5}{*}{LLaMA}
& Full-attn         & 31.44 & 25.07 & 29.40 & 16.89 & 17.00 & 11.79 & 34.22 & 23.25 & 26.69 & 72.50 & 91.65 & 43.74 & 5.95 & 98.20 & 54.04 & 51.49 & 39.58 \\
& StreamingLLM      & 21.27 & 23.48 & 24.05 & 14.26 & 13.43 & 8.46  & 33.47 & 22.28 & 26.76 & 66.50 & 90.32 & \textbf{44.46} & \textbf{7.26} & 38.24 & 54.55 & 52.56 & 33.83 \\
& Vertical\_Slash   & 20.87 & \textbf{24.54} & 26.19 & 17.12 & 14.37 & 8.38  & 32.84 & 22.33 & 26.85 & 63.50 & \textbf{91.38} & 44.12 & 0.98 & \textbf{98.61} & 54.22 & 36.41 & 36.48 \\
& FlexPrefill       & \textbf{28.31} & 23.79 & 28.78 & \textbf{19.24} & 16.22 & 10.58 & 33.60 & \textbf{22.95} & \textbf{27.06} & 70.50 & 90.74 & 43.81 & 1.37 & 77.50 & 54.23 & \textbf{54.09} & 36.66 \\
\rowcolor{gray!10}
& Ours              & 27.79 & 23.82 & \textbf{28.86} & 16.29 & \textbf{16.84} & \textbf{11.74} & \textbf{34.50} & 22.94 & 27.01 & \textbf{72.50} & 90.67 & 43.82 & 3.53 & 96.92 & \textbf{54.72} & 49.65 & \textbf{38.23} \\
\midrule
\multirow{5}{*}{Qwen}
& Full-attn         & 11.53 & 13.99 & 31.83 & 10.88 & 10.02 & 7.12  & 32.52 & 20.65 & 22.58 & 71.50 & 89.47 & 46.68 & 3.92 & 98.42 & 59.63 & 66.57 & 37.33 \\
& StreamingLLM      & 11.70 & 13.68 & 31.39 & 11.34 & 9.77 & 5.94  & 32.63 & 19.85 & \textbf{22.52} & 72.00 & 89.02 & 45.76 & \textbf{4.18} & 73.83 & 59.22 & 65.28 & 35.51 \\
& Vertical\_Slash   & 10.70 & 13.40 & 31.59 & 11.30 & 9.87 & \textbf{8.06}  & \textbf{32.70} & 20.65 & 22.47 & 70.50 & \textbf{89.73} & \textbf{46.00} & 3.46 & 94.25 & \textbf{60.21} & \textbf{66.36} & 36.51 \\
& FlexPrefill       & 8.73  & 13.91 & 29.96 & \textbf{11.36} & 8.76 & 6.69  & 32.16 & \textbf{21.08} & 22.37 & 70.50 & 88.29 & 45.66 & 2.03 & 71.67 & 58.94 & 60.68 & 34.90 \\
\rowcolor{gray!10}
& Ours              & \textbf{14.57} & \textbf{14.23} & \textbf{32.18} & 10.73 & \textbf{9.93} & 7.24  & 32.21 & 20.76 & 22.46 & \textbf{72.50} & 89.05 & 45.69 & 3.99 & \textbf{94.58} & 59.28 & 65.27 & \textbf{37.17} \\
\bottomrule
\end{tabular}
}
\caption{Accuracy (\%) of different attention mechanisms across models on LongBench.}
\label{tab:longbench_all_models}
\end{table*}

\begin{table*}[h]
\centering
\small
\begin{tabular}{llcccccc|c}
\toprule
\textbf{Models} & \textbf{Methods} & \textbf{4k} & \textbf{8k} & \textbf{16k} & \textbf{32k} & \textbf{64k} & \textbf{128k} & \textbf{Avg} \\
\midrule
\multirow{4}{*}{LLaMA} 
& Full-attn        & 95.67 & 93.75 & 93.03 & 87.26 & 84.37 & 78.13 & 88.70\\
& Streaming LLM    & \textbf{96.62} & 92.06 & 84.54 & 66.77 & 46.69 & 37.03 & 70.61\\
&Vertical\_Slash   & 95.81 & 92.82 & 93.26 & 88.96 & \textbf{85.09} & 58.18 & 85.69\\
&FlexPrefill       & 95.46 & 93.18 &  93.53 & \textbf{90.02} & 84.73 & \textbf{75.03}  & \textbf{88.66}\\
\rowcolor{gray!10}
& Ours             & 95.98 & \textbf{93.27} & \textbf{93.67} &  87.79 & 84.53 & 74.91 & 88.36\\
\midrule
\multirow{5}{*}{Qwen} 
& Full-attn        & 94.92 & 93.01 & 92.31 & 86.54 & 66.76 & 22.72  & 76.04 \\
& Streaming LLM    & 93.74 & 90.91 & 74.39 & 57.81 & 25.48 & 15.88 & 59.70 \\
& Vertical\_Slash  & 94.91 & 92.16 & \textbf{92.17} & \textbf{85.59} & 60.10 & 24.78  & 74.95 \\
& FlexPrefill      & 93.04 & 90.69 & 90.16 & 80.37 & 40.42 & 25.43 & 70.01 \\
\rowcolor{gray!10}
& Ours             & \textbf{94.98} & \textbf{92.86} & 89.74 & 84.68 & \textbf{66.79} & \textbf{25.71} & \textbf{75.79} \\
\bottomrule
\end{tabular}
\caption{Accuracy (\%) on RULER benchmark across models of different attention mechanisms.}
\label{tab:ruler_all}
\end{table*}

\subsection{Difference-aware Stripe Sparsity Identification}
\label{sec:diff_sparse}

Numerous prior studies \cite{zhang2023ho,yang2024pyramidinferpyramidkvcache,li2024snapkv} have observed significant column-wise correlation in attention score distributions—namely, a small subset of keys consistently receives high attention across multiple consecutive queries. However, as discussed in Section~\ref{sec:attention_diversity}, these column-wise correlations are not always stable or effective, often exhibiting vanishing and reappearing behaviors. This observation motivates our strategy to focus on global information by identifying the corresponding keys and values for each query segment individually, rather than determining a set of global key-value pairs based on a subset of queries\cite{jiang2024minference10acceleratingprefilling,lai2025flexprefillcontextawaresparseattention}.

As discussed in Section~\ref{sec:sparsity_strategies}, to efficiently identify these coordinates, we compress queries through block-average compression and compute their dot product with all keys. The result is directly compared to the average anchor value \(\mathrm{avgpool}(\boldsymbol{x}_{\text{a}})\) from Equation~\ref{eq:anchor} through numerical difference. By setting a hyperparameter \(\theta\), we compute only the discrete keys and values whose difference is below this threshold. This approach outperforms static top-k strategies, achieving performance consistent with dynamic top-cdf strategies while avoiding costly sorting operations.

We define the sparsity mask as:
\[
\text{mask} = \mathbb{I} \left( 
    \mathrm{avgpool}(\boldsymbol{x}_{\text{a}}) - \frac{\mathrm{avgpool}(Q) K^{\top}}{\sqrt{d}} \leq \theta 
\right)
\]

\begin{equation}
\mathcal{S} = \left\{ (i, j) \mid \text{mask}(i, j) = 1 \right\}
\label{eq:sparsity_set}
\end{equation}
where \(\boldsymbol{x}_{\text{a}}\) is the approximate highest attention score, \(\mathrm{avgpool}(\boldsymbol{x}_{\text{a}})\) is its pooled average, \(\mathrm{avgpool}(Q)\) is the pooled queries, \(\theta\) is the comparison threshold, \(\text{mask} \in \{0, 1\}^{n \times m}\) is the binary mask, \(\mathcal{S}\) is the set of coordinates to be activated, and \(\mathbb{I}(\cdot)\) is the indicator function. The detailed implementation is provided in Algorithm~\ref{alg:stripe_sparse}.

\subsection{Fine-Grained Sparse Computation}
\label{sec:kernel_optimized}

In contrast to prior strategies that load contiguous key-value blocks, our fine-grained sparse computation methodology selectively loads multiple discrete key-value pairs based on discrete key-value coordinates. Throughout the computational process, we adhere to the sharding strategy of FlashAttention, employing the same computation logic. However, compared to block-sparse approaches, our discrete key-value loading, as discussed in Section~\ref{sec:Granularities}, achieves a higher sparsity rate due to lower granularity with negligible additional overhead, thereby significantly enhancing the efficiency of sparse computation.

To formalize the fine-grained sparse computation, we construct the reduced key and value sets by discretely loading key-value pairs based on the sparse coordinate set \(\mathcal{S}\) from Equation~\ref{eq:sparsity_set}. The index set \(\mathcal{I}\) is defined as:
\begin{equation}
\mathcal{I} = \{ j \mid (i, j) \in \mathcal{S} \},
\label{eq:index_set}
\end{equation}
and the reduced key and value sets are constructed as:
\begin{equation*}
K' = \mathrm{load\_discrete}(K, \mathcal{S})
\end{equation*}
\begin{equation}
V' = \mathrm{load\_discrete}(V, \mathcal{S})
\label{eq:reduced_kv}
\end{equation}
where $\mathrm{load\_discrete}(M, \mathcal{S}) = \{ M[j, :] \mid j \in \mathcal{I} \}$ denotes selecting the key or value rows from the matrix $M$ (e.g., $K$ or $V$) corresponding to the indices in $\mathcal{I}$. The sparse attention output is then computed as:
\begin{equation}
\text{Output} = A(Q, K', V')
\label{eq:sparse_output}
\end{equation}
where \(A(Q, K', V')\) denotes the attention computation, with the granularity of key-value loading modified from contiguous blocks (as in FlashAttention\cite{dao2022flashattention}) to discrete keys and values based on the coordinates in \(\mathcal{S}\). The detailed implementation is provided in Algorithm~\ref{alg:sparse_comp}.
\subsection{Kernel Optimization and Algorithm}
To further accelerate sparse attention computation, we apply kernel-level optimizations to all algorithms, with the goal of satisfying two objectives simultaneously: (1) maximizing parallel computation capacity and (2) introducing no additional memory overhead. To achieve this, we introduce an additional hyperparameter \( \textit{step} \), which enables the simultaneous identification of coordinates corresponding to \( \textit{step} \) query blocks. If any of these blocks contain a key that satisfies the condition defined in Equation~\ref{eq:sparsity_set}, all \( \textit{step} \) consecutive blocks are marked as active for computation, allowing unified processing and enhanced parallelism. Meanwhile, to avoid redundant overhead, we temporarily cache the intermediate results generated in Section~\ref{sec:anchor_computation}, and reuse them in Section~\ref{sec:kernel_optimized}. This design maximizes computational efficiency while introducing only negligible memory overhead compared to the original key-value cache. The complete implementation is detailed in Algorithms~\ref{alg:anchor_comp}, \ref{alg:stripe_sparse}, and \ref{alg:sparse_comp}.

\begin{figure*}[ht]
    \centering
    \begin{subfigure}[b]{0.322\textwidth}
        \centering
        \includegraphics[width=\textwidth]{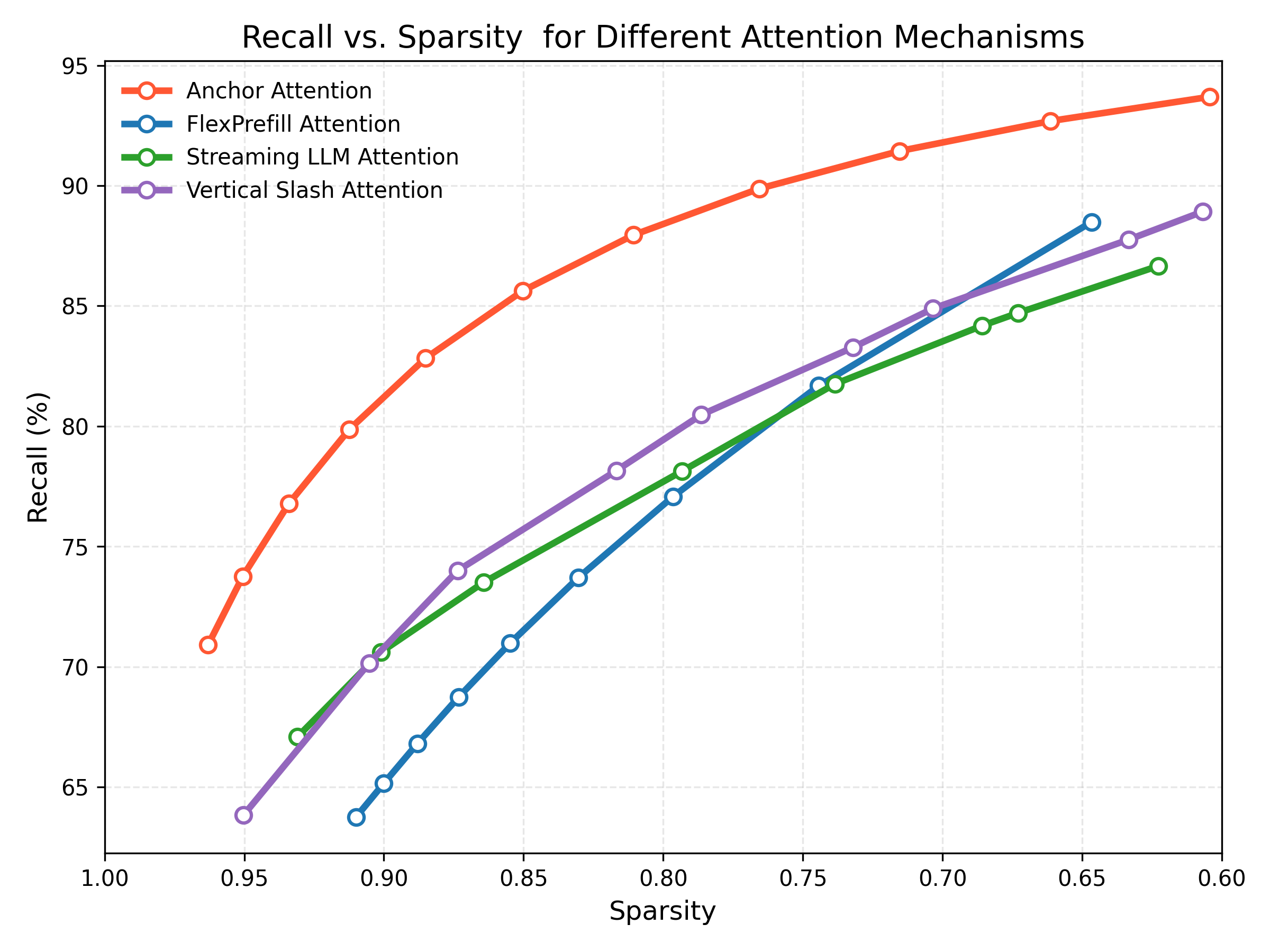}
        \caption{Recall vs. Sparsity}
        \label{fig:recall_sparsity}
    \end{subfigure}
    \hfill
    \begin{subfigure}[b]{0.322\textwidth}
        \centering
        \includegraphics[width=\textwidth]{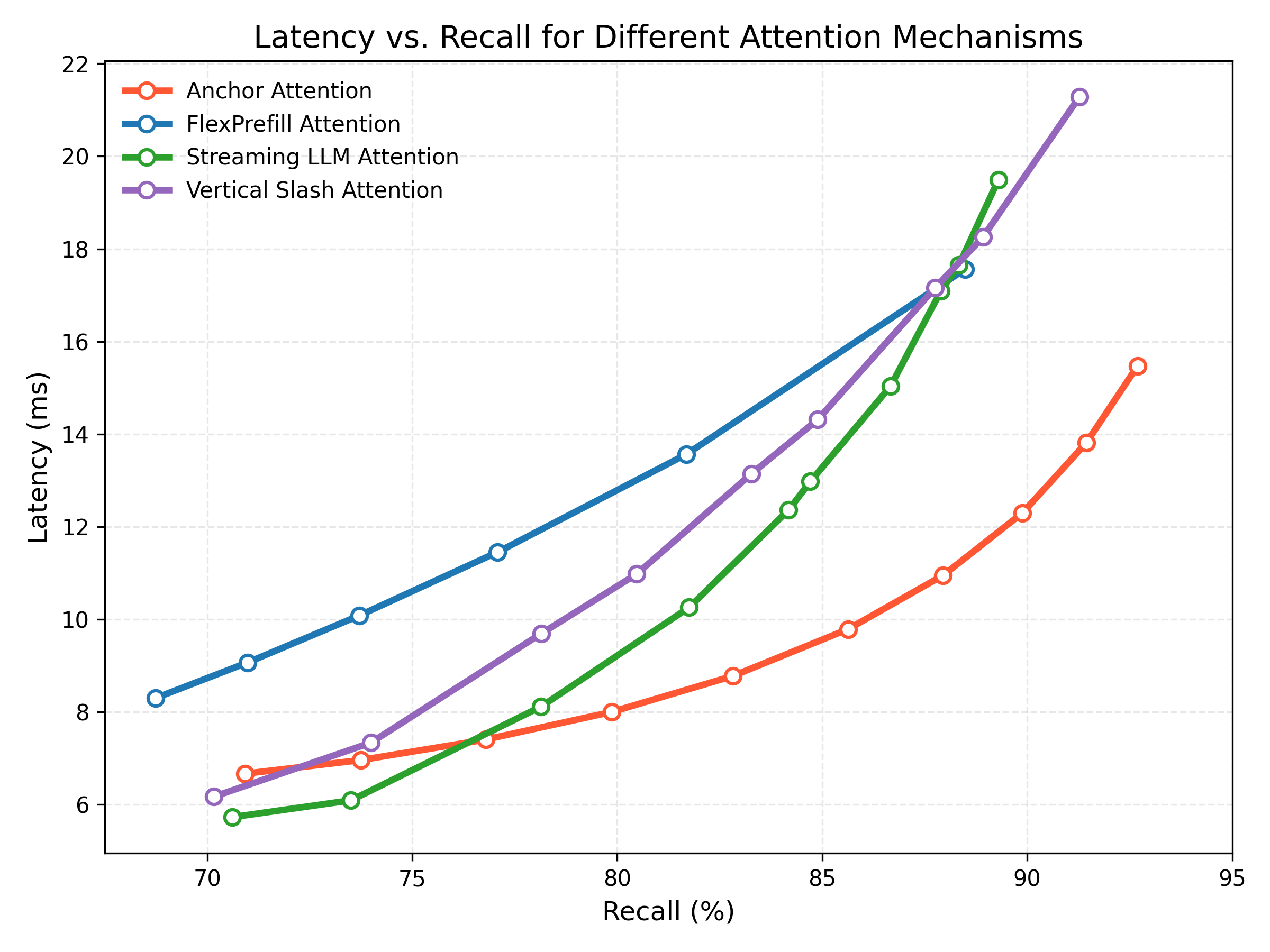}
        \caption{Latency vs. Recall}
        \label{fig:time_recall}
    \end{subfigure}
    \hfill
    \begin{subfigure}[b]{0.322\textwidth}
        \centering
        \includegraphics[width=\textwidth]{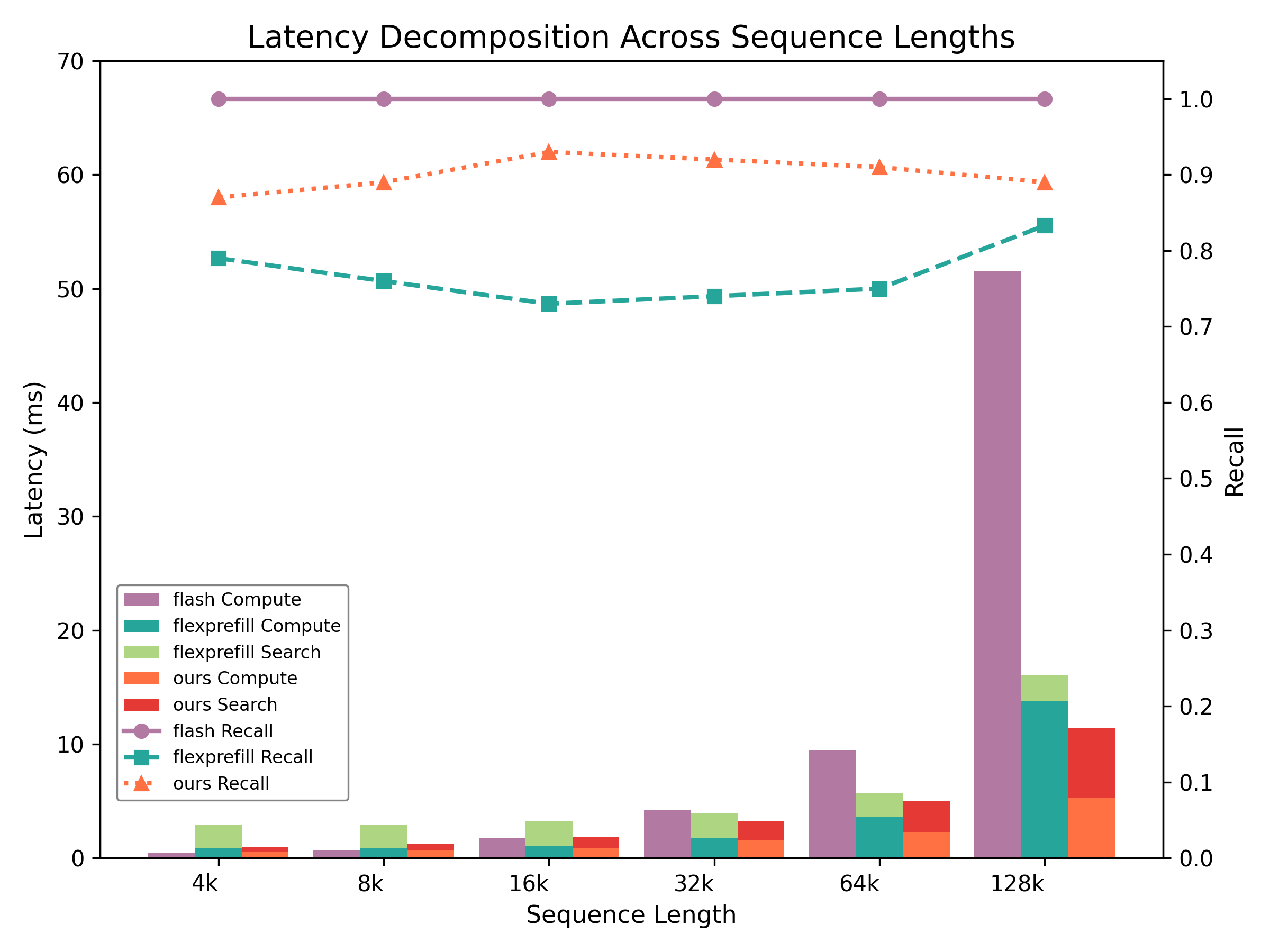}
        \caption{Latency vs. Length}
        \label{fig:method_time}
    \end{subfigure}
    \caption{Performance metrics for recall, sparsity, and efficiency across different methods. Figures (a), (b), and (c) are all generated using data from the Ruler benchmark. Specifically, Figures (a) and (b) correspond to the 128k length. For Figure (b), the latency metric is defined as the average computation time per head.}
    \label{fig:performance_metrics}
\end{figure*}

\section{Experiment}
\label{sec:experiment}
\subsection{Setup}

\textbf{Models} Our evaluation is conducted on two advanced large language models (LLMs) that natively support up to 128K context length in their pre-trained form:
(i) \textbf{LLaMA-3.1-8B-Instruct}~\cite{touvron2023llama},
(ii) \textbf{Qwen2.5-7B-Instruct}~\cite{qwen2025qwen25technicalreport}.
Both models are evaluated in their pre-trained form without fine-tuning, ensuring a fair and consistent comparison.


\textbf{Benchmark} We evaluate models on three representative long-context benchmarks, each designed to test different aspects of long-context understanding and retrieval:
(i) \textbf{LongBench}~\cite{bai2024longbenchbilingualmultitaskbenchmark}, a multilingual, multi-task benchmark covering question answering, summarization, classification, and retrieval, with diverse input formats;
(ii) \textbf{RULER}~\cite{hsieh2024rulerwhatsrealcontext}, a synthetic benchmark that enables controlled variations in context length and reasoning complexity, including tasks such as multi-hop tracing and aggregation;
(iii) \textbf{Needle-in-a-Haystack}~\cite{kamradt2023llmtest}, a stress test designed to evaluate accurate retrieval performance in ultra-long contexts.

\textbf{Baseline} We evaluate four baselines for accelerating prefill attention:  
(i) \textbf{Full-attn}, dense attention implemented via FlashAttention~\cite{dao2022flashattention};  
(ii) \textbf{Vertical\_Slash}~\cite{jiang2024minference10acceleratingprefilling}, which selects a fixed set of important vertical and slash positions;  
(iii) \textbf{StreamingLLM}~\cite{xiao2024efficientstreaminglanguagemodels}, retaining only key tokens from initial and local window regions;  
(iv) \textbf{FlexPrefill}~\cite{lai2025flexprefillcontextawaresparseattention}, a dynamic method selecting attention blocks based on top-cdf scoring, representing recent state-of-the-art.

\textbf{Implementation} All experiments are conducted on a single NVIDIA A100 GPU with 80GB memory, leveraging Triton~\cite{10.1145/3315508.3329973} for optimized GPU computations. To ensure fair comparison, all methods adopt a uniform block size of 128. Across all datasets, our method and FlexPrefill use consistent hyperparameter settings: for ours, we set $\theta = 12$ and $\text{step} = 16$; for FlexPrefill, we use $\gamma = 0.95$, $\tau = 0.1$, and $\text{min\_budget} = 1024$.
For LongBench, which has relatively shorter average sequence lengths, StreamingLLM uses a global window and a local window of 1024, and Vertical\_Slash sets both vertical and slash window sizes to 1024. For other datasets, StreamingLLM adopts a global window of 1024 and a local window of 8192, while Vertical\_Slash uses a vertical window of 1024 and a slash window of 8192. In the latency-recall evaluation, we uniformly choose to generate data using the ruler and report the averaged results.

\subsection{Result}
\textbf{Longbench} To demonstrate the applicability of our method to nearly all input scenarios, we selected the LongBench benchmark for accuracy evaluation. LongBench encompasses a variety of tasks that exhibit input diversity, testing whether our method maintains high accuracy across different inputs. The accuracy results are presented in Table~\ref{tab:longbench_all_models}.

\textbf{Ruler} To demonstrate the potential of our approach for large language models handling varying context lengths, we conducted evaluations on multiple methods using the ruler benchmark. Table~\ref{tab:ruler_all} shows that, as context length increases, our method consistently maintains accuracy close to that of full KV computations.

\begin{figure}[htbp]
    \centering
    \begin{subfigure}[t]{0.49\columnwidth}
        \centering
        \includegraphics[width=\columnwidth]{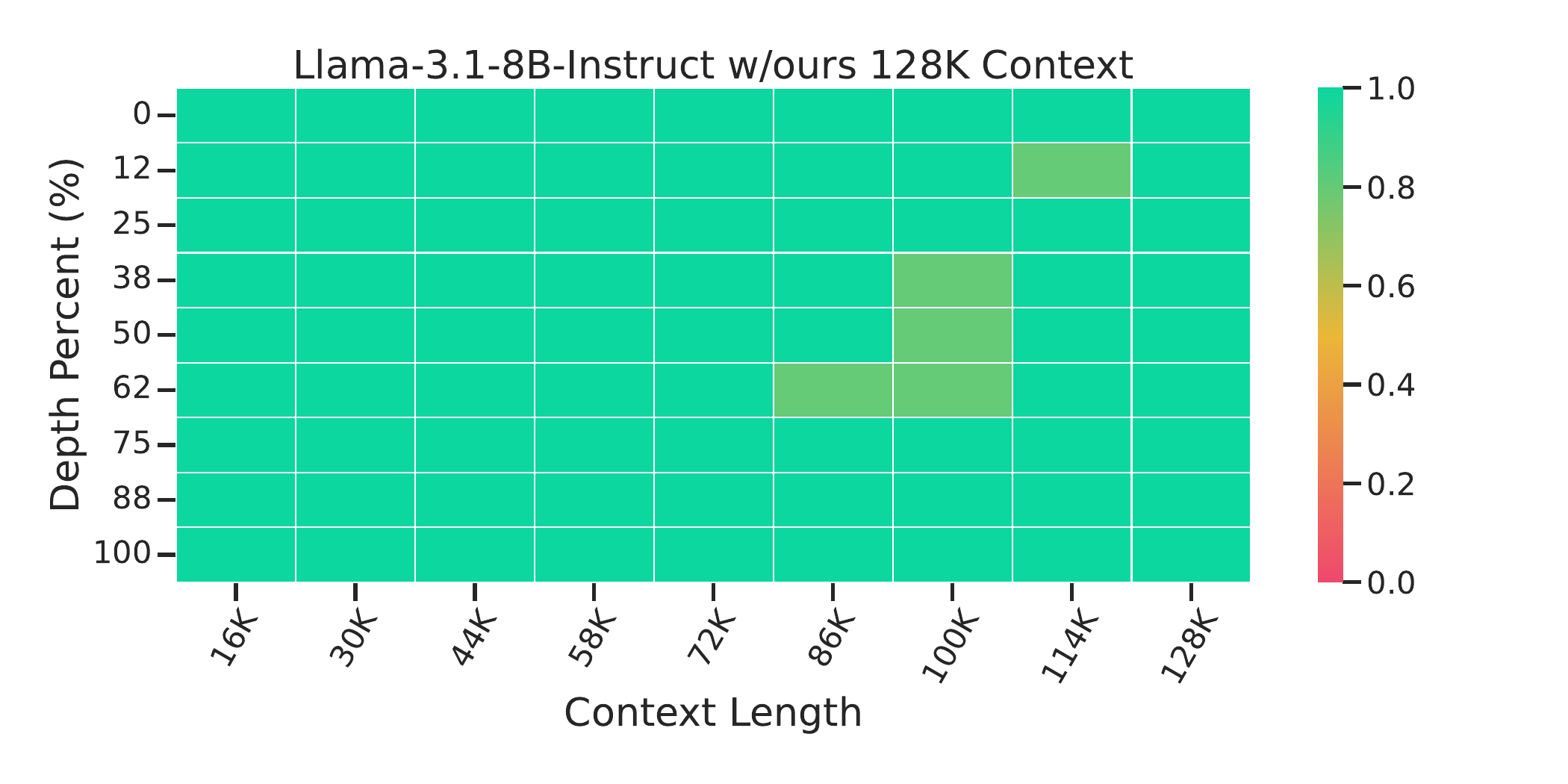}
        \label{fig:ours}
    \end{subfigure}
    \hspace{-\columnsep}
    \begin{subfigure}[t]{0.49\columnwidth}
        \centering
        \includegraphics[width=\columnwidth]{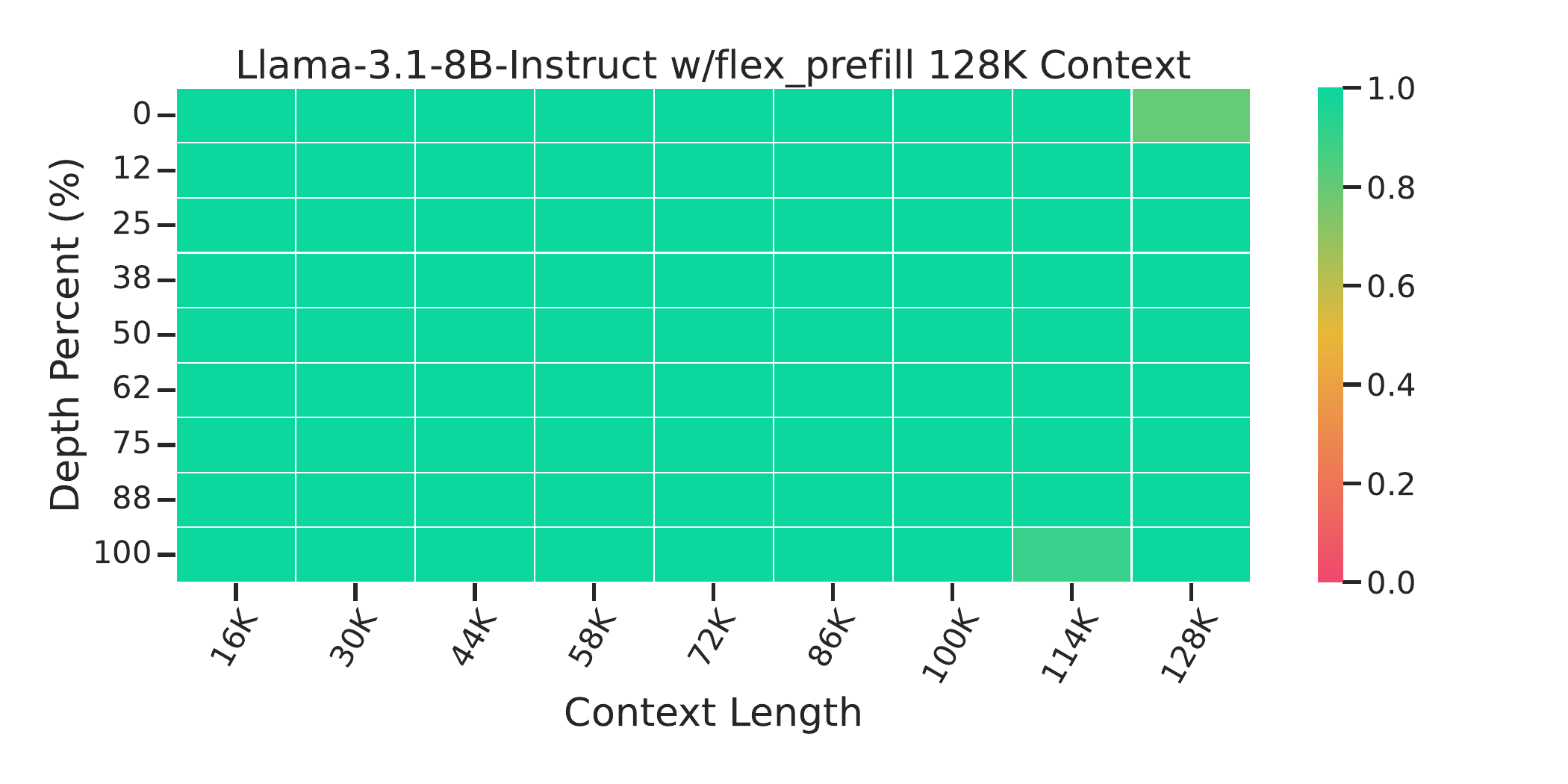}
        \label{fig:flex_prefill}
    \end{subfigure}

    \vspace{-\baselineskip}

    \begin{subfigure}[t]{0.49\columnwidth}
        \centering
        \includegraphics[width=\columnwidth]{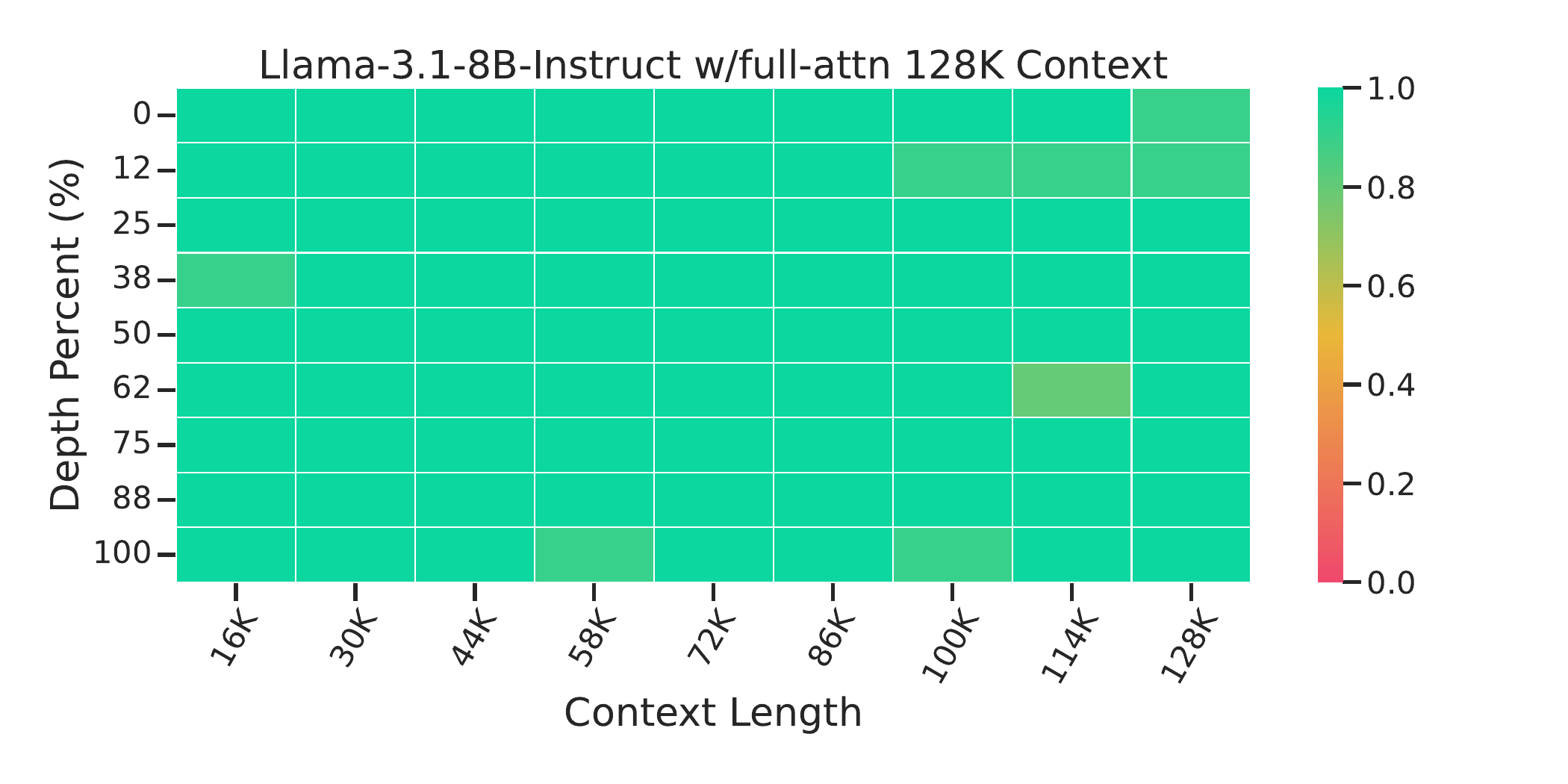}
        \label{fig:full_attn}
    \end{subfigure}
    \hspace{-\columnsep}
    \begin{subfigure}[t]{0.49\columnwidth}
        \centering
        \includegraphics[width=\columnwidth]{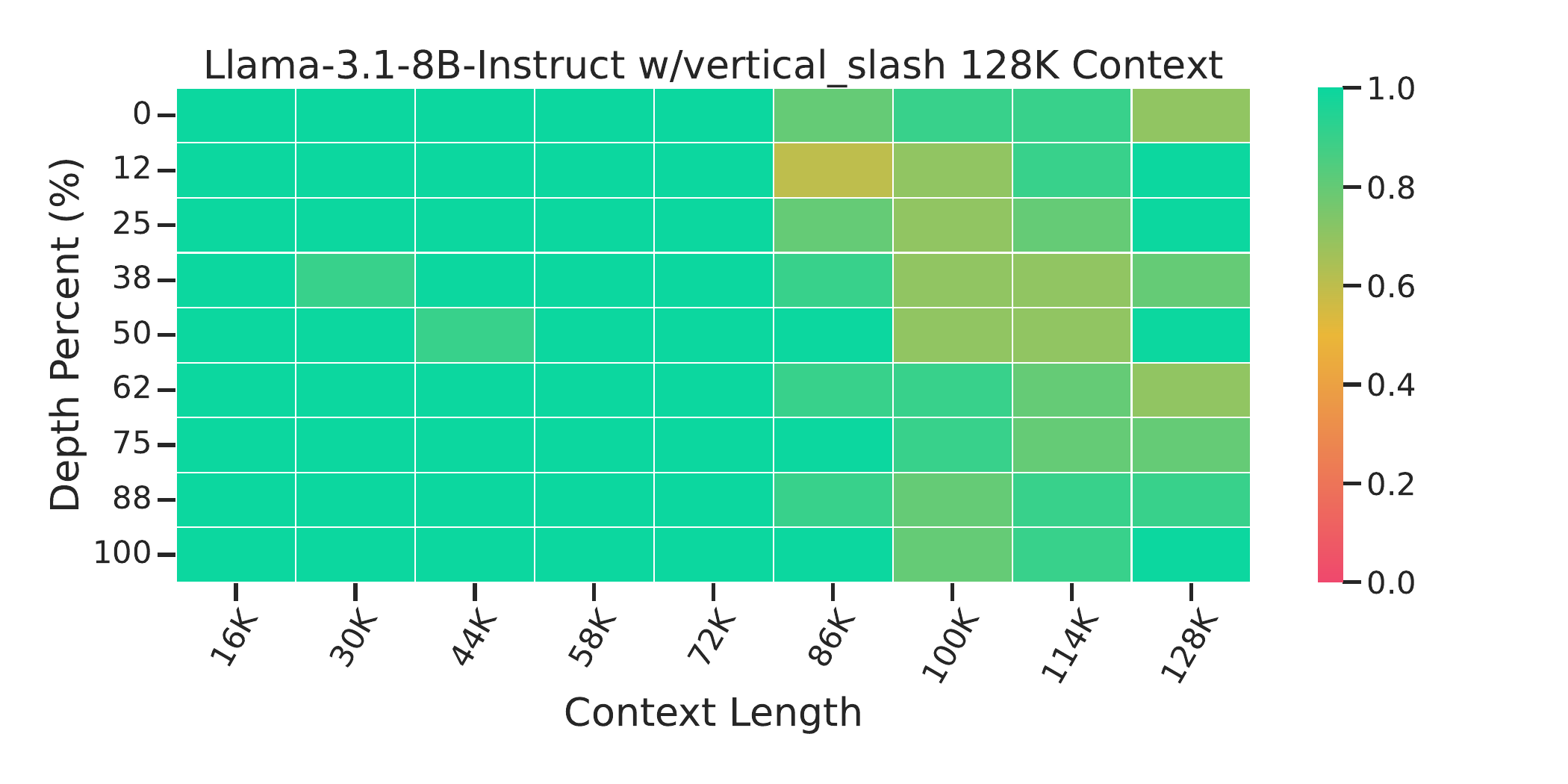}
        \label{fig:vertical_slash}
    \end{subfigure}

    \caption{Comparison of attention patterns on Needle-in-a-Haystack tasks (128K context).}
    \label{fig:needle_comparison}
\end{figure}

\textbf{Needle-in-a-Haystack} 
As shown in Figure~\ref{fig:needle_comparison}, we present the results of the Needle-in-a-Haystack task across different context lengths and depth percentages. The results indicate that both our method and FlexPrefill can dynamically adapt the sparsity rate based on input variations, achieving performance comparable to full attention. In contrast, the static strategy Vertical\_Slash shows a noticeable accuracy drop as the context length increases.

\textbf{Recall vs. Sparsity}
\label{sec:Performance_Latency} We adjust the hyperparameters of different methods to obtain varying sparsity rates and compare the recall performance of different strategies under each sparsity level. As shown in Figure~\ref{fig:recall_sparsity}, our method achieves the highest sparsity rate under the same recall level.

\textbf{Latency vs. Recall} Prior work primarily differs in search strategies, with distinctions arising from the blocks requiring computation. Our method abandons block-level sparsity strategies, instead adopting a finer-grained computation strategy that loads multiple discrete keys and values at once. 
As illustrated in Figure~\ref{fig:time_recall}, at the same recall level, our strategy significantly outperforms other methods in terms of time efficiency.

\textbf{Latency vs. Length} \label{sec:Search_Time}   
Compared to prior strategies, our approach considers the entire region during search. This higher search overhead also brings us more accurate recognition, which is reflected in the recall curves and the computation time section. As shown in Figure~\ref{fig:method_time}, our method incurs additional recognition time in most cases, but it achieves a higher important recognition ratio, thereby optimizing overall time efficiency and recall. 

\subsection{Ablation Study}
\begin{table}[h]
  \centering
  \footnotesize                      
  \renewcommand{\arraystretch}{1.0}
  \setlength{\tabcolsep}{4pt}          
  \resizebox{\linewidth}{!}{%
    \begin{tabular}{lcccc}
      \toprule
      \textbf{Anchor Attention} & \textbf{$\theta$} & \textbf{Sparsity (\%)} & \textbf{Recall (\%)} & \textbf{Time (ms)} \\
      \midrule
      \multirow[c]{6}{*}{With Anchor}
          & 10.0 & 97\% & 70.9 & 5.7 \\
          & 11.0 & 93\% & 76.8 & 6.4 \\
          & 12.0 & 89\% & 82.8 & 8.2 \\
          & 13.0 & 81\% & 88.0 & 10.9 \\
          & 14.0 & 72\% & 91.4 & 13.8 \\
          & 15.0 & 61\% & 94.7 & 19.3 \\
      \midrule
      \multirow[c]{6}{*}{Without Anchor}
          & 10.0 & 83\% & 69.5 & 9.3 \\
          & 11.0 & 69\% & 83.7 & 14.6 \\
          & 12.0 & 52\% & 90.2 & 29.5 \\
          & 13.0 & 47\% & 95.8 & 41.3 \\
          & 14.0 & 18\% & 96.2 & 49.7 \\
          & 15.0 & 3\% & 98.5 & 57.2 \\
      \bottomrule
    \end{tabular}%
  }
  \caption{Ablation study of Anchor Attention. Results are averaged over all heads using a 128k context length.}
  \label{tab:anchor_attention_ablation}
\end{table}

\textbf{Anchor Importance} In this section, we assess the impact of introducing anchors when searching for important tokens by comparing sparsity, recall, and computation time under different values of \textbf{$\theta$}. As shown in Table~\ref{tab:anchor_attention_ablation}. The original attention(With Anchor) consistently achieves high recall rates while maintaining impressively low sparsity, indicating effective attention guidance. In contrast, the Without Anchor configuration, which sets the anchor as a zero tensor in implementation, requires significantly higher sparsity to reach comparable recall levels. This suggests that fixed thresholding alone, without anchor guidance, is less adept at capturing the global attention distribution efficiently, resulting in a less optimal sparsity-recall balance.

\section{Related Work}

\textbf{LLM Inference Acceleration}  
Inference acceleration techniques aim to reduce the latency and memory overhead of large language models (LLMs) during text generation.  
At the system level, FlashAttention~\cite{dao2022flashattention} significantly improves attention computation efficiency by optimizing memory access patterns,  
while RingAttention~\cite{liu2023ringattentionblockwisetransformers} distributes attention workloads across multiple devices to achieve parallel acceleration.  
PagedAttention~\cite{kwon2023efficientmemorymanagementlarge} further enhances overall inference performance through efficient KV cache management.

\textbf{Sparse Attention}  
The quadratic complexity of attention has driven extensive research into sparse attention strategies to improve the inference efficiency of large language models (LLMs).  
Importantly, attention distributions in LLMs are inherently sparse—many attention weights are close to zero and can be safely pruned without significantly affecting model performance~\cite{child2019generatinglongsequencessparse}.    
More recent methods such as H2O~\cite{zhang2023ho} and SnapKV~\cite{li2024snapkv} prune unimportant tokens by comparing cumulative attention scores.  
Although partially effective, these methods offer limited acceleration benefits during the prefill stage.StreamingLLM~\cite{xiao2024efficientstreaminglanguagemodels} significantly improves efficiency by retaining only initial and recent tokens,  
but often misses critical information from intermediate regions.  
MInference~\cite{jiang2024minference10acceleratingprefilling} accelerates the prefill stage by applying statically determined attention patterns,  
but such static designs are often suboptimal for diverse and dynamic inputs.  
FlexPrefill~\cite{lai2025flexprefillcontextawaresparseattention} improves adaptivity via runtime-driven dynamic pattern selection,  
yet relies heavily on local information, limiting its ability to capture globally important positions. Recently, research has shifted toward building general-purpose sparse attention frameworks rather than designing architectures tailored specifically to LLM characteristics.  For example, SpargeAttn\cite{zhang2025spargeattnaccuratesparseattention} leverages similarity-based filtering and quantization to accelerate attention,  
while X-Attention\cite{xu2025xattentionblocksparseattention} introduces an antidiagonal scoring mechanism to efficiently prune irrelevant blocks.  
Furthermore, most existing methods rely on block-level granularity, where block size fundamentally constrains the achievable sparsity ceiling.  
Therefore, there is an urgent need for a lower-granularity sparse attention mechanism with a stronger emphasis on global context, in order to mitigate the increasingly heavy computational burden during the prefill stage as context lengths continue to grow.

\section{Conclusion}
In this work, we propose \textbf{AnchorAttention}, a difference-aware, dynamic sparse attention mechanism designed to address the computational challenges faced by Large Language Models (LLMs) during the prefill phase under long-context settings. The method efficiently identifies critical attention regions at a finer \textit{stripe-level} granularity.

To further improve speed, we implement all operators at the kernel level. By combining pattern-based anchor computation, difference-aware stripe sparsity identification, and fine-grained sparse computation, \textbf{AnchorAttention} achieves higher sparsity and superior computational efficiency compared to existing methods. At a sequence length of 128k, it achieves a 1.44$\times$ speedup while maintaining a higher recall rate.

\section*{Limitations}
Our evaluation is limited to the LLaMA-3.1-8B-instruct and Qwen2.5-7B-instruct models, and we have not yet validated the generality of AnchorAttention across a broader range of architectures and model scales; future work will extend our analysis to additional models.
Additionally, we do not account for the importance of slash and row-wise patterns, as our design prioritizes maximizing parallelism while ensuring high recall rates.
Furthermore, this work focuses exclusively on the prefill phase of attention computation and does not analyze the impact or adaptivity of our method during the decode phase; subsequent studies will investigate performance and sparsity behavior during generation.

\section*{Ethics Statement}
We believe this work raises no ethical concerns. Attention is a key component in Transformers, widely used in Large Language Models (LLMs). Therefore, accelerating the execution of attention is beneficial for developing LLM applications that address diverse societal challenges.

\bibliography{emnlp2023}
\bibliographystyle{acl_natbib}

\onecolumn
\newpage
\appendix
\section{Sparsity Heatmap Comparison}
\label{app:Difference_method_sparisty}

\begin{figure*}[htbp]
    \centering

    \begin{subfigure}[t]{0.32\textwidth}
        \centering
        \includegraphics[width=\textwidth]{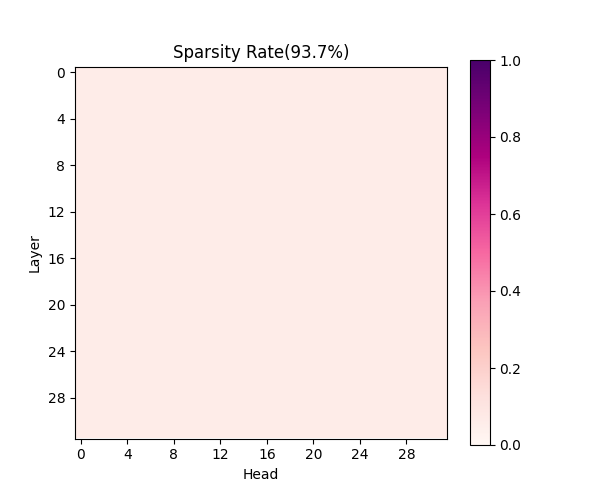}
        \caption{Top-K (4096)}
        \label{fig:heat_map_top_k_app}
    \end{subfigure}
    \hfill
    \begin{subfigure}[t]{0.32\textwidth}
        \centering
        \includegraphics[width=\textwidth]{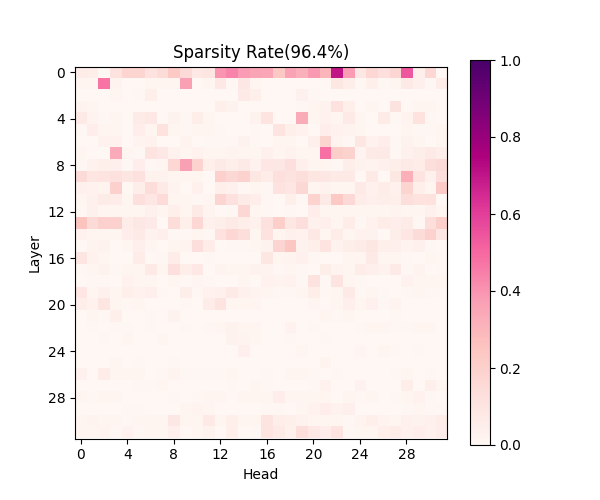}
        \caption{Top-CDF (0.95)}
        \label{fig:heat_map_top_cdf_app}
    \end{subfigure}
    \hfill
    \begin{subfigure}[t]{0.32\textwidth}
        \centering
        \includegraphics[width=\textwidth]{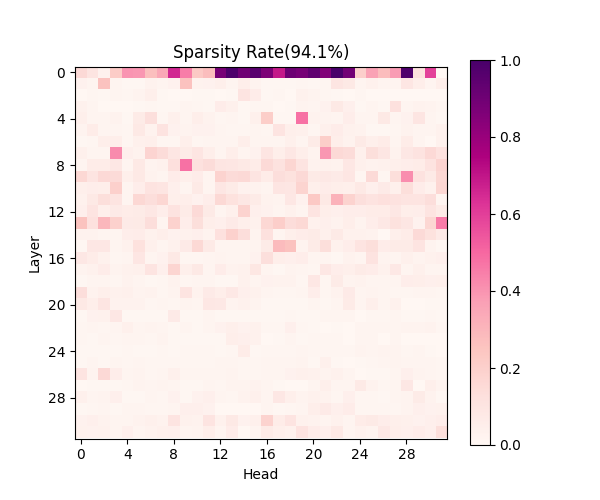}
        \caption{Difference-Aware (11)}
        \label{fig:heat_map_difference_app}
    \end{subfigure}
    \caption{Sparsity heatmaps under different sparsity strategies. The recall heatmap corresponds to Figure~\ref{fig:heatmaps}.}
    \label{fig:heatmaps_sparist_n1}
\end{figure*}

Figure~\ref{fig:heatmaps} presents the per-layer, per-head recall distributions on the LLaMA-3.1-8B-instruct model using the 128k ruler datasets. In Figure~\ref{fig:heatmaps_sparist_n1}, we further visualize the sparsity levels achieved under this target recall for different identification strategies. The results indicate that our proposed Difference-Aware strategy achieves sparsity patterns comparable to those of top-cdf while maintaining similar recall performance.

\section{Dynamic Sparsity Heatmap}
\label{app:dynamic_method_sparisty}

\begin{figure*}[htbp]
    \centering
    \begin{subfigure}[t]{0.32\textwidth}
        \centering
        \includegraphics[width=\textwidth]{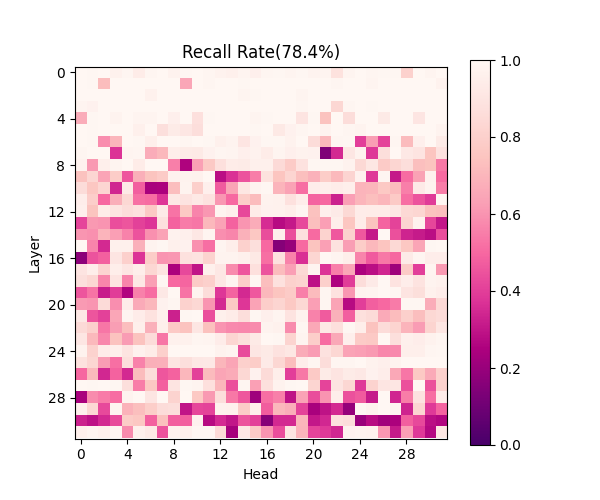}
        \caption{Top-K (4096) }
        \label{fig:heatmap_recall_topk}
    \end{subfigure}
    \hfill
    \begin{subfigure}[t]{0.32\textwidth}
        \centering
        \includegraphics[width=\textwidth]{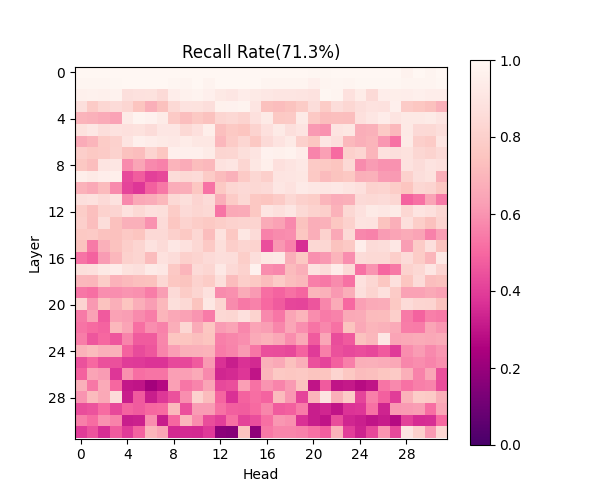}
        \caption{Top-CDF (0.95) }
        \label{fig:heatmap_recall_topcdf}
    \end{subfigure}
    \hfill
    \begin{subfigure}[t]{0.32\textwidth}
        \centering
        \includegraphics[width=\textwidth]{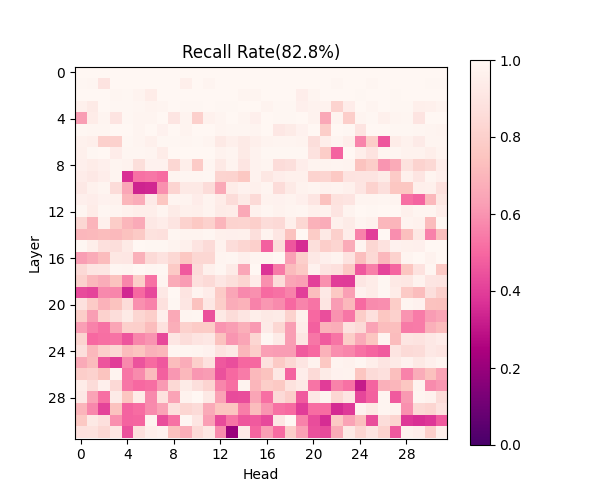}
        \caption{Difference-Aware (11) }
        \label{fig:heatmap_recall_diffaware}
    \end{subfigure}
    \caption{Recall heatmaps under different sparsity identification strategies.}
    \label{fig:heatmaps_recall_n2}
\end{figure*}

\begin{figure*}[htbp]
    \centering

    \begin{subfigure}[t]{0.32\textwidth}
        \centering
        \includegraphics[width=\textwidth]{fig/sparse_topk_attention__top_k_4096__sparsity.png}
        \caption{Top-K (4096)}
        \label{fig:heat_map_top_k_app}
    \end{subfigure}
    \hfill
    \begin{subfigure}[t]{0.32\textwidth}
        \centering
        \includegraphics[width=\textwidth]{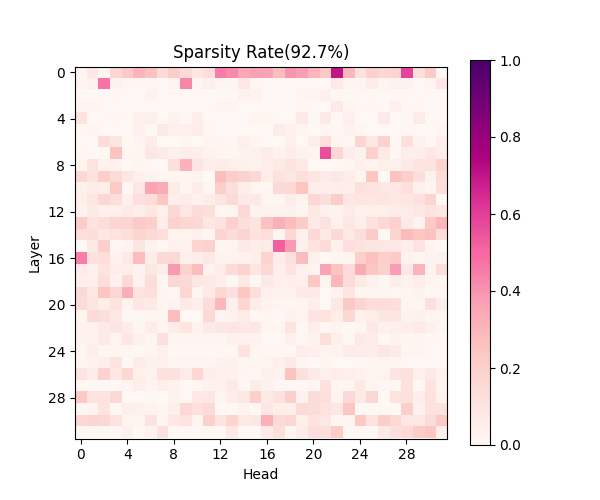}
        \caption{Top-CDF (0.95)}
        \label{fig:heat_map_top_cdf_app}
    \end{subfigure}
    \hfill
    \begin{subfigure}[t]{0.32\textwidth}
        \centering
        \includegraphics[width=\textwidth]{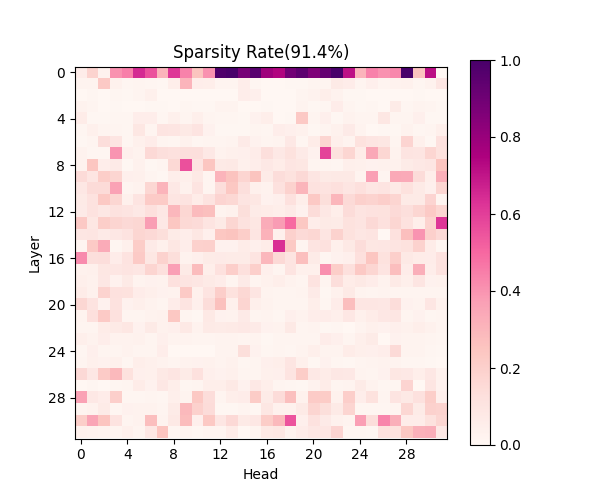}
        \caption{Difference-Aware (11)}
        \label{fig:heat_map_difference_app}
    \end{subfigure}
    \caption{Sparsity heatmaps for different sparsity strategies.}
    \label{fig:heatmaps_app_n2}
\end{figure*}

To demonstrate the dynamic nature of the heatmap, we selected a distinct dataset with the same length of 128k. The recall rates under different sparsity strategies are shown in Figure~\ref{fig:heatmaps_recall_n2}, with the corresponding sparsity rates depicted in Figure~\ref{fig:heatmaps_app_n2}. It is evident that, as the input changes, both the top-cdf and difference-aware methods can effectively capture variations in sparsity rates.

\section{Algorithm}

We provide the complete pseudocode of our proposed sparse attention inference pipeline, consisting of three key stages:

\paragraph{Algorithm~\ref{alg:anchor_comp}: Anchor Computation.}
This algorithm performs efficient block-wise attention to obtain an approximate estimation of the attention result, which is used later for sparsity identification. The query matrix $Q$ is divided into blocks $Q_i$ and interacts only with a small number of key-value blocks (e.g., the initial block and a local window). The accumulated attention values $\text{Acc}_i$, normalization terms $L_i$, and maximum logits $M_i$ are computed and cached. These intermediate results are reused in the final sparse attention step to avoid redundant computation.

\paragraph{Algorithm~\ref{alg:stripe_sparse}: Stripe Sparsity Identification.}
Based on the averaged queries and approximated attention output from the previous step, this algorithm identifies informative positions through a lightweight thresholding mechanism. By comparing the approximated anchor score $x_a$ with new attention estimates, it selects positions with scores close to the anchor. This enables the construction of stripe-wise sparse indices $F\_idx$ without computing full attention maps, greatly improving efficiency.

\paragraph{Algorithm~\ref{alg:sparse_comp}: Sparse Attention Computation.}
This stage computes the final attention output using only the key/value blocks selected via sparse indexing. For each query block $Q_i$, the algorithm loads its corresponding anchor values ($M_i$, $L_i$, $\text{Acc}_i$) and incrementally accumulates the attention using the sparse key-value entries. This computation avoids redundant processing and yields high sparsity while maintaining high recall and accuracy.

\floatstyle{plain}
\begin{algorithm*}
\small
\caption{Anchor Computation}
\label{alg:anchor_comp}
\begin{algorithmic}[1]
\REQUIRE $Q, K, V \in \mathbb{R}^{N \times d}$ (FP16), block sizes $b_q$, $b_{kv}$, step size $step$
\STATE Divide $Q$ into $T_m = N / b_q$ blocks $\{Q_i\}$; $K$, $V$ into $T_n = N / b_{kv}$ blocks $\{K_j\}, \{V_j\}$
\FOR{$i = 1$ to $T_m$}
    \STATE Load $Q_i$, $K_1$, $V_1$ into shared memory
    \STATE Compute initial attention: $qk \leftarrow Q_i \cdot K_1$
    \STATE $m \leftarrow \max(qk, \text{dim}=-1)$
    \STATE $p \leftarrow \exp(qk - m)$, $l \leftarrow \sum(p, \text{dim}=-1)$, $acc \leftarrow p \cdot V_1$
    \STATE Determine local window range:
    \STATE \hspace{1em} $j_{\text{start}} \leftarrow \max(2, \lfloor (i - 1) / \text{step} \rfloor \cdot \text{step} \cdot (b_q / b_{kv}))$
    \STATE \hspace{1em} $j_{\text{end}} \leftarrow i \cdot (b_q / b_{kv})$
    \FOR{$j = j_{\text{start}}$ to $j_{\text{end}}$}
        \STATE Load $K_j$, $V_j$ into shared memory
        \STATE Compute $qk \leftarrow Q_i \cdot K_j$, $m' \leftarrow \max(m, \max(qk))$
        \STATE $p \leftarrow \exp(qk - m')$, $\alpha \leftarrow \exp(m - m')$
        \STATE $l \leftarrow l \cdot \alpha + \sum(p)$, $acc \leftarrow acc \cdot \alpha + p \cdot V_j$
        \STATE Update $m \leftarrow m'$
    \ENDFOR
    \STATE Write $M_i \leftarrow m$, $L_i \leftarrow l$, $Acc_i \leftarrow acc$
\ENDFOR
\RETURN $M$, $L$, $Acc$
\end{algorithmic}
\end{algorithm*}

\begin{algorithm*}
\small
\caption{Stripe Sparsity Identification}
\label{alg:stripe_sparse}
\begin{algorithmic}[1]
\REQUIRE $Q, K \in \mathbb{R}^{N \times d}$ (FP16), anchor score $Acc$, block sizes $b_q$, $b_{kv}$, threshold $\theta$, step size $step$
\STATE Compute averaged query $Q_{\text{mean}} \leftarrow \text{avgpool}(Q, b_q)$
\STATE Compute anchor average $x_a \leftarrow \text{avgpool}(Acc, b_q)$
\STATE Divide $Q_{\text{mean}}$ into $T_m = N / (b_q \cdot \text{step})$ blocks $\{Q^m_i\}$
\STATE Divide $K$ into $T_n = N / b_{kv}$ blocks $\{K_j\}$
\FOR{$i = 1$ to $T_m$}
    \STATE Initialize $f_c \leftarrow 0$, $f_{\text{idx}} \leftarrow \emptyset$
    \STATE $j_{\text{end}} \leftarrow (i - 1) \cdot \text{step} \cdot (b_q / b_{kv})$
    \FOR{$j = 2$ to $j_{\text{end}}$}
        \STATE Load $K_j$
        \STATE Compute $qk \leftarrow Q^m_i \cdot K_j$
        \STATE $\text{mask} \leftarrow (x_a - qk) < \theta$
        \STATE Append matching indices to $f_{\text{idx}}$, count to $f_c$
    \ENDFOR
    \STATE Write $F_{\text{idx}}^{(i)} \leftarrow f_{\text{idx}}, \quad F_c^{(i)} \leftarrow f_c$
\ENDFOR
\RETURN $F_{\text{idx}}, F_c$
\end{algorithmic}
\end{algorithm*}

\begin{algorithm*}
\small
\caption{Sparse Attention Computation (Reusing Anchor and Stripe Outputs)}
\label{alg:sparse_comp}
\begin{algorithmic}[1]
\REQUIRE Query $Q$, Key $K$, Value $V \in \mathbb{R}^{N \times d}$ (FP16), precomputed $M$, $L$, $Acc$ (from Alg.~\ref{alg:anchor_comp}), and sparse indices $F_{\text{idx}}, F_c$ (from Alg.~\ref{alg:stripe_sparse}); block sizes $b_q$, $b_{kv}$; step size $step$
\STATE Divide $Q$ into $T_m = N / b_q$ blocks $\{Q_i\}$
\STATE Divide $M$, $L$, $Acc$ into $\{M_i\}$, $\{L_i\}$, $\{Acc_i\}$
\STATE Divide $F_c$, $F_{\text{idx}}$ into $\{F_c^{(k)}\}$, $\{F_{\text{idx}}^{(k)}\}$ where $k = \lfloor (i-1) / \text{step} \rfloor$

\FOR{$i = 1$ to $T_m$}
    \STATE Load $Q_i$, and corresponding $M_i$, $L_i$, $Acc_i$
    \STATE Initialize $m \leftarrow M_i$, $l \leftarrow L_i$, $acc \leftarrow Acc_i$
    \STATE Let $k \leftarrow \lfloor (i - 1) / \text{step} \rfloor$
    \STATE  \textcolor{blue}{\textit{\# Simultaneously load multiple discrete coordinate chunks from $F_{\text{idx}}^{(k)}$}}
    \FOR{\textbf{each} index chunk $f_{\text{idx}}^j$ in $F_{\text{idx}}^{(k)}$}
        \STATE Load sparse key/value: $K_j = K[f_{\text{idx}}^j]$, $V_j = V[f_{\text{idx}}^j]$
        \STATE Compute $qk = Q_i \cdot K_j$, $m' = \max(m, \max(qk))$
        \STATE $p = \exp(qk - m')$, $\alpha = \exp(m - m')$
        \STATE $l = l \cdot \alpha + \sum(p)$, $acc = acc \cdot \alpha + p \cdot V_j$
        \STATE Update $m = m'$
    \ENDFOR
    \STATE Write output $O_i = acc / l$
\ENDFOR
\RETURN Final attention output $O$
\end{algorithmic}
\end{algorithm*}

\end{document}